\documentclass[runningheads]{llncs}
\usepackage{graphicx}
\usepackage{hyperref}
\usepackage{amsfonts}
\begin{document}
\title{Knowledge Graph Completion using Structural and Textual Embeddings\thanks{Accepted in AIAI 2024 - 20th International Conference on Artificial Intelligence Applications and Innovations}}

%
%
\author{Sakher Khalil Alqaaidi\inst{1} \and
Krzysztof Kochut\inst{1}}
%
%
\institute{School of Computing, University of Georgia
\\
\email{\{sakher.a,kkochut\}@uga.edu}\\
}
\maketitle              
\begin{abstract}

Knowledge Graphs (KGs) are widely employed in artificial intelligence applications, such as question-answering and recommendation systems. However, KGs are frequently found to be incomplete. While much of the existing literature focuses on predicting missing nodes for given incomplete KG triples, there remains an opportunity to complete KGs by exploring relations between existing nodes, a task known as relation prediction. In this study, we propose a relations prediction model that harnesses both textual and structural information within KGs. Our approach integrates walks-based embeddings with language model embeddings to effectively represent nodes. We demonstrate that our model achieves competitive results in the relation prediction task when evaluated on a widely used dataset.

\end{abstract}
\section{Introduction} \label{sec:introduction}
Knowledge Graphs (KGs) store informational facts in a structured format of connected triples. A triple consists of a semantic relation that binds a head (subject) entity node with a tail (object) entity node. KGs have been widely employed in several artificial intelligence applications on a global data scale, such as question-answering \cite{yani2021challenges} and recommendation systems \cite{sheu2020context}. Due to their enormous sizes \cite{hu2020open,bollacker2008freebase}, Knowledge Graphs suffer from being incomplete \cite{xu2016knowledge}.
Given that the KG incompleteness problem affects its utilization, several works proposed Knowledge Graph Completion (KGC) models. These works tackled three types of KGC, that are Link Prediction (LP), Relation Prediction (RP), and Triple Classification (TC). The LP task aims at finding the missing head or tail node in a triple. The RP task aims at finding the valid relations for a given pair of head and tail nodes. The TC task aims at determining the plausibility of a complete triple.

After surveying the literature, we found that much of the literature works focused on the LP task, whereas the RP task remained a secondary objective in some models \cite{shen2022comprehensive}. A few models proposed an RP variant based on a link prediction-oriented design \cite{lin2015modeling}. However, we see that the RP task holds equal importance to the LP task and can effectively contribute to KGC. That is by identifying all the possible relations between node pairs. For instance, if \textit{Apple} and \textit{California} are two nodes in a knowledge graph and connected using the relation \textit{has\_store\_in}, we could be interested in exploring other possible relations between them. For example, one relation that can be established between them is \textit{headquarters\_in} after identifying \textit{Apple} as the node for the technological company.

Recently, KGC models have achieved remarkable accomplishments by representing graphs in the form of multi-dimensional float vectors \cite{shen2022comprehensive}. Therefore, KGC models can be categorized into three groups according to the representation type used. First, KGC models utilized the graph's structural information \cite{bordes2013translating,sun2019rotate}. Second, KGC models utilized the nodes' meta information \cite{yao2019kg}. Third, KGC models utilized a combination of the two information types \cite{xu2016knowledge,wang2021structure,youn2022kglm}. In terms of meta information utilization, several models have used the textual content within node entity names as input for language models, enabling the generation of representation embeddings. This approach has demonstrated efficacy, particularly in inductive settings \cite{daza2021inductive}, that is when prediction is conducted for nodes not seen during model training. Notably, Masked Language Models (MLMs) showed astonishing results in the KGC task using nodes' text. However, these language models suffer from a costly fine-tuning phase in terms of computational resource usage, such as time and memory. In contrast, Pre-trained Language Models (PLMs) have significantly lower computational load and still provide meaningful representation of words without training the model again.

Although language models cap present powerful text representation, employing the graph structural information is still crucial in the KGC task \cite{wang2021structure}, primarily due to the entity ambiguity problem encountered when encoding text content. Particularly, because language models cannot represent an entity efficiently relying on short text. Similar to the previous \textit{Apple} and \textit{California} example, language model embeddings for the \textit{Apple} node cannot lead to determining whether it is the technological company or just the fruit without having additional clues other than the node name. Otherwise, the relation \textit{planted\_in} would be predicted because \textit{Apple} could also be a node for the fruit. Indeed, determining the node identity would be possible only if the text content is detailed as in this sentence: \textit{Apple is a technology company headquartered in California}. However, such detailed description is not usually provided in KGs without employing connections to external knowledge resources. In contrast, the node's structural details can support revealing the node's identity depending on topological information, such as the neighborhood and the walks obtained from there \cite{perozzi2014deepwalk}.



To this end, we propose a \textbf{r}elations \textbf{p}rediction model (RPEST) that utilizes the \textbf{e}mbeddings of \textbf{s}tructural and \textbf{t}extual features in knowledge graphs. Our model employs a walk-based graph structure algorithm, namely Node2Vec \cite{grover2016node2vec}, to replace the language model fine-tuning step with structural embeddings training. Additionally, we exploit pre-trained language models efficiently to capture text contextualized representation and to avoid the costly fine-tuning phase in MLMs. We show that our choice of language model has significantly lower overhead compared to masked language models. Notably, our model achieves competitive results in the relation prediction task when compared to the best accomplishments in Freebase, a widely used benchmark. Furthermore, we analyze the effectiveness of our model's components in a dense ablation study. To the best of our knowledge, this is the first work that considers textual embeddings along with graph walks-based embeddings for the RP task. The code to reproduce the results is available online \footnote{https://github.com/sa5r/KGRP}

\section{Background and Related Work} \label{sec:related_work}

Given a knowledge graph $\mathcal{G} = \{ \mathcal{E} , \mathcal{R} , \mathcal{T} \}$, where $\mathcal{E}$ is a set of entity nodes, $\mathcal{R}$ is a set of relations of size $k$, and $\mathcal{T}$ is a set of entities text. $\mathcal{G}$ is constructed of facts; each fact is represented as a directed triple $(h, r, t)$, where $h,t \in \mathcal{E}$, $h$ is a head entity node, $t$ is a tail entity node, $r \in \mathcal{R}$, and $r$ is a relation with a direction from $h$ to $t$; the direction is essential to hold the fact, thus $(t, r, h)$ is not equivalent to $(h, r, t)$. The relation prediction task targets finding the probability of all relations in $\mathcal{R}$ for the given nodes $h$ and $t$, formally:

\begin{equation}
 \phi(h, t) = Pr(\mathcal{R})   
\end{equation}

where $\phi$ is a probability values scoring function. In contrast, the link prediction task aims to find a missing head node $(?, r, t)$ or a missing tail node $(h,r,?)$ in a triple. Recent achievements in knowledge graph completion have followed a representation learning approach \cite{chen2020knowledge}. Specifically, models targeted learning a function that can efficiently represent knowledge graph triples in a multi-dimensional float values space $\mathbb{R}$, i.e., continuous vector embeddings. KGC models can be categorized into methods based on the node's structural information, textual content information, or a combination of both.

\subsection{Graph Structural Approaches}

Several translation-based methods have achieved remarkable results in the link prediction task \cite{sun2019rotate,bordes2013translating,lin2015learning}. TransE \cite{bordes2013translating} started a translation-based stream of methods for unsupervised graph embeddings learning. The translation technique uses distance-based scoring function to find close embedding values of head and tail nodes given the relation between them. Specifically, finding the minimum value of the embeddings of the three triple elements in this equation $|h + r - t|$, where $h, r,$ and $t$ are represented in a d-dimensional space $\mathbb{R}^d$, i.e., $h, r, t \in \mathbb{R}^d$. Since TransE used a unified space for nodes and relations. TransR \cite{lin2015learning} proposed a multi-relational spaces model to tackle the problem of nodes involved in multiple facts in the knowledge graph. Formally, each relation is represented in the space $\mathbb{R}^k$, and the knowledge graph facts are represented in the space $\mathbb{R}^{k \times d}$. However, this approach came at the cost of additional computation due to the extended spaces. TaRP is a relation prediction model \cite{cui2021type}. TaRP employed different translation-based methods such as TransE and RotateE, to generate representation embeddings. Additionally, TaRP incorporated the node type identification and encoding to boost the relation prediction results. A different course to generate node embeddings is using neural networks. Shallom \cite{demir2021shallow} proposed a shallow neural network layers for the relation prediction task.

\subsection{Textual Content Approaches}

The textual content in KGs has been exploited in different ways. The node's long description helped in predicting the relations for nodes that were not seen during model training, i.e., inductive settings \cite{tagawa2019relation}. However, retrieving each node's text description requires additional steps and the employment of external knowledge bases.

BERT \cite{devlin2018bert} emerged as a masked language model with a state of the art performance on a variety of Natural Language Processing (NLP) tasks KG-BERT \cite{yao2019kg} was one of the first models to utilize BERT for the KGC task; KG-BERT targeted mainly the LP task. Additionally, two variants were proposed for the RP and TC tasks. Inspired by this work, BERTRL \cite{zha2022inductive} linearized the triples around each relation and trained the model to have better inductive performance. Nevertheless, language model-based approaches suffered from serious issues. BERT and its successors of masked language models \cite{liu2019roberta}, incorporate a costly fine-tuning phase in terms of time and memory. In large language models (LLMs) such as GPT and Llama, the computation needs are worst \cite{touvron2023llama,radford2019language}.

In contrast, pre-trained language models (PLMs) have significant lower computational load. Furthermore, PLMs provide meaningful representation of text words for new datasets without retraining the model. For instance, Glove \cite{pennington2014glove} is a PLM that provides multi-dimensional vectors representation for a single word with $O(n)$ complexity, where $n$ is the PLM's vocabulary size. However, Glove operates at the word level, unlike MLMs and LLMs that can represent text sequences as contextualized embeddings. Text sequences can consist of mutli-word terms or sentences. A node's text content can include multiple words, such as `Alfred Nobel'.

\subsection{Hybrid Approaches}

Here we review KGC models that employed a hybrid approach based on textual and structural representation. LASS model \cite{shen2022joint} employed BERT's fine-tuning for the textual encoding and used the resulted loss values to construct the semantic embeddings. A variant for triple classification was proposed in the paper. StAR model \cite{wang2021structure} targeted handling the overwhelming performance in language models by reusing the graph elements' embeddings. However, the model achieved a satisfactory performance only after combining the its textual representation with an existing graph embeddings, such as RotateE. The KGLM model \cite{youn2022kglm} re-ran the original Roberta language model training with additional text generated from translating KG triples into normal sentences. The model added an entity/relation-type embedding layer to include the graph structure in training. In KEPLER \cite{wang2021kepler}, authors jointly optimized the language model training objective with the structural training objective for the link prediction task.

Some models achieved better results in the link prediction task through trying different negative sampling (contrastive learning) approaches in the text embeddings training \cite{wang2022simkgc,he2023mocosa}. The SimKGC model \cite{wang2022simkgc} incorporated the structured information by re-ranking candidates based on path distances between target nodes in the graph, whereas the MoCoSA model \cite{he2023mocosa} used the independently trained structural embeddings and fused them with the textual embeddings.


\begin{figure}[t]
\centering
\includegraphics[width=0.75\textwidth]{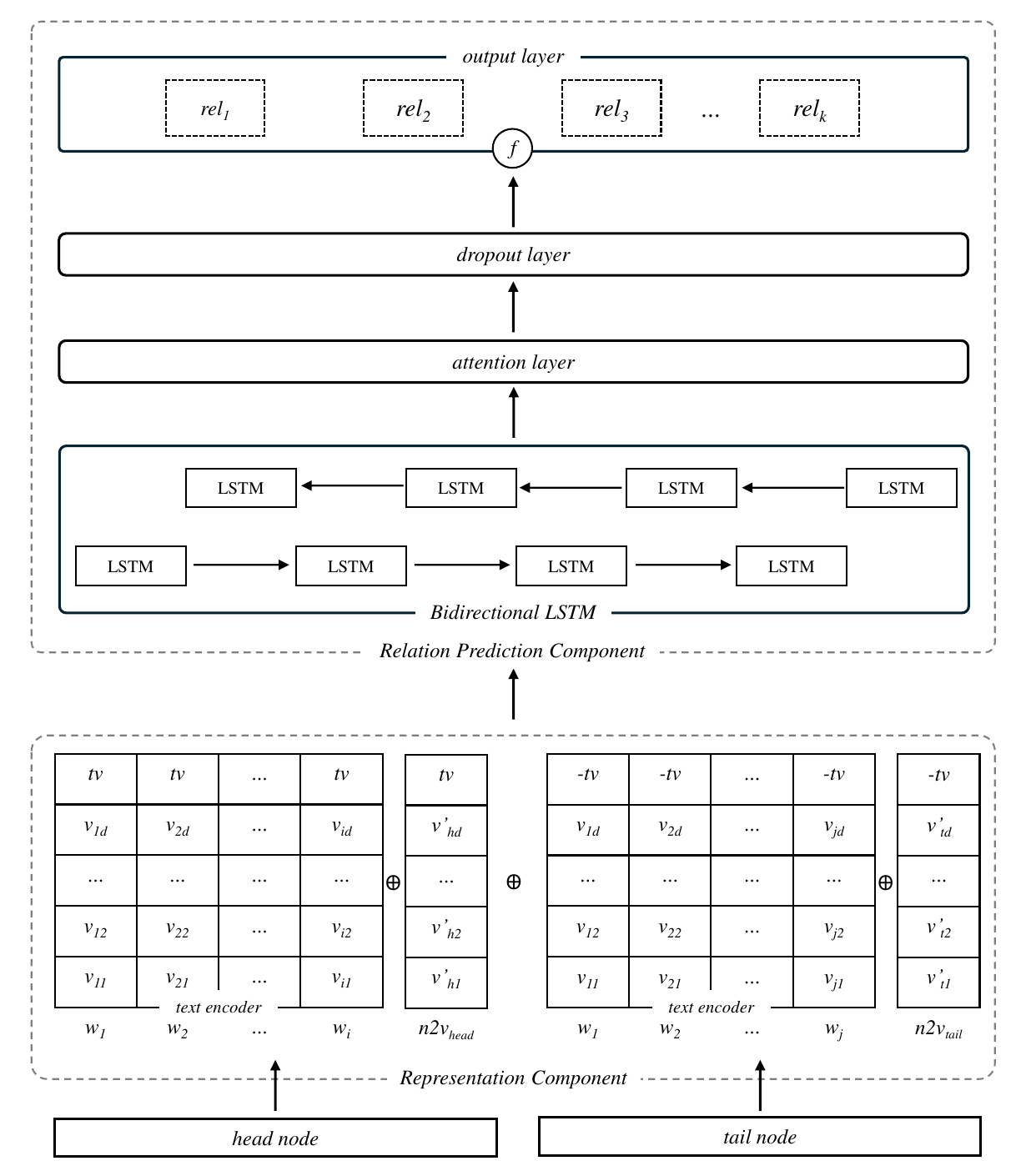}
\caption{Our model's diagram showing the node representation component and the relations prediction component.}
\label{fig:diagram}
\end{figure}

\section{Methodology} \label{sec:methodology}

Our model aims to find the probability of every predefined relation that can establish a fact by connecting a head node to a tail node. Formally, $Pr(\mathcal{R}|h,t)$ where $\mathcal{R}$ is a set of predefined relations, $h$ is the head node, and $t$ is the tail node. Our model takes as an input a combination of the node structural representation and the node textual representation. Our model conducts a supervised learning for a neural network to predict the relations' probabilities. Accordingly, our model consists of two main components, the representation component and the relations prediction component. The former incorporates our approach to generate the structural representation and the textual representation, whereas the later incorporates a recurrent neural network and the prediction layer. Figure \ref{fig:diagram} shows the architecture of our model. 

\subsection{Structural Representation Training}

We incorporate the nodes structural information using the Node2Vec \cite{grover2016node2vec} model. Node2Vec has been known for its efficiency and for being versatile. The model is an unsupervised learning algorithm for graph features representation. Training in Node2Vec aims at optimizing a function that maximizes the log-probability of observing a neighborhood for a node, based on the neighborhood nodes' representation as the following:
\[ \max\limits_{f} \sum_{u \in V} log \, Pr(N_{s}(u) | f(u)) \]
where $V$ is the nodes set in a graph $\mathcal{G}$, $N_{s}(u)$ is the neighborhood for a node $u$, and $f(u)$ is the features representation for the node $u$. The source node neighborhood is constructed from sampling nodes identified following biased random walk strategies. The strategies could be either Breadth-First Sampling (BFS) or Depth-First Sampling (DFS). In the former, the sampled nodes are adjacent to the source node, whereas in the later, the sampled nodes are found while taking further steps in the walk paths, so it is not necessary for the nodes to be immediately connected to the source node, but a direct path should lead to the source node.

Node2Vec is used in our model as a preliminary step before training the prediction neural network, that is described in section \ref{sec:nn}, this initial phase replaces the fine-tuning step in MLMs, which is not needed in our implementation as described in section \ref{sec:text_encoder}. After training the Node2Vec algorithm on the graph, we get a representation $v^{'}$ of $d$ vectors for each node as shown in Figure \ref{fig:diagram}.

\subsection{Text Encoding} \label{sec:text_encoder}

We employ Glove \cite{pennington2014glove} pre-trained word embeddings to represent the nodes' textual content. Glove language model is an unsupervised learning algorithm trained on a word-word co-occurrence probabilities using a global scale text corpus. The training objective in Glove is minimizing the difference between two words co-occurrence as the following:

\[
    J = \sum^{V}_{i,j = 1} f(X_{ij})(w^{T}_{i}\tilde{w_j} + b_i + \tilde{b_j} - log\,X_{ij})^2
\]

where $V$ is the number of words in Glove's vocabulary, $i$ and $j$ are identification numbers for two words, $w_i$ and $b_i$ are the vector and the bias for word $i$, $w_j$ and $b_j$ are the vector and bias for word $j$, $X_{ij}$ is the number of times word $i$ occurred in the context of word $j$, and $f$ is a weighting function for the co-occurrences.

We chose Glove due to it is superiority among other language models, such as the skip-gram and continuous bag-of-words (CBOW) \cite{mikolov2013efficient}, and for providing immediate text representation without fine-tuning. However, Glove works at the word level, accordingly, the pre-trained embeddings $E^{d}$ represent each word in Glove's vocabulary with vectors of size $d$ dimensions. In our implementation, Glove encodes each node as $n \times d$ two-dimensional vector matrix, where n is the number of words to be considered in each node. When the number of words in a node is greater than $n$, we omit the words of index $(n + 1)$ and greater. In contrast, for nodes with number of words below $n$, we append embeddings of zeros to keep the embeddings consistent; each padding embedding has $d$ vectors.


KG node names usually consist of multiple words. For instance, in Freebase \cite{bollacker2008freebase}, the average entity name length is 2.7 words. Although multi-word terms and sentences can be represented efficiently in MLMs and LLMs, PLMs still can present a contextualized encoding by employing a Recurrent Neural Network (RNN). Consequently, we employ a bidirectional long-short term memory (LSTM) layer in our model's neural network. RNNs are capable of encoding the latent features in text sequences \cite{sak2014long}. For the Out-of-Vocabulary (OOV) words problem \cite{zhuang2023out}, we find the longest possible match for a word in Glove's vocabulary or take the word letter embeddings average in the worst case. However, OOV words still occupy less than 2\% for most datasets when using Glove, due to its large vocabulary.

We enhance the recognition of the relation direction in the input by appending additional vector at the top of each node's representation, this approach has been proven to help predicting the proper relations based on the direction \cite{alqaaidi2023multiple}. The head node uses a fixed value in the added vector and the tail node uses the same value but multiplied by -1.

\subsection{Relation Prediction} \label{sec:nn}

The architecture of our model's neural network ensures the utilization of the latent features in the combined node embeddings.
The input embeddings for each pair of nodes consist of a head node representation concatenated with a tail node representation. Each node's representation consists of the textual embeddings concatenated with the structural embeddings as shown in Figure \ref{fig:diagram}. For each pair of nodes, the bidirectional LSTM layer generates new contextualized embeddings, that are forwarded to an attention layer to selectively focus on certain parts of the input data, allowing the model to weigh different inputs differently based on their relevance to the task \cite{vaswani2017attention}. In the following dropout layer \cite{hinton2012improving} we enhance the generalization performance. Ultimately, the output layer uses the Sigmoid function \cite{han1995influence} to give the probability of each relation. Our model computes the neural network training loss using the cross entropy loss function.

\section{Experiments} \label{sec:experiments}

We evaluated our model on FB15K, a subset of the FreeBase dataset \cite{bollacker2008freebase}. FreeBase has been maintained by Google for several years and has been widely used for KG-related tasks; its content includes data collected from several sources such as Wikipedia. The last release of the complete dataset holds about 1.9 billion triples \footnote{\textit{https://developers.google.com/freebase/}}. The FB15K subset holds 1,345 relations and 14,951 entities. The number of triples in the training portion is 483,142, the validation portion has 50,000 triples, and the testing portion has 59,071 triples.

We used PyTorch to train our model. The used hardware consisted of NVIDIA A100-SXM-80GB GPU node, a main memory of 50GB, and AMD EPYC MILAN (3rd gen) CPU processor; we utilized 8 cores of the CPU. Table \ref{tab:training} shows our training settings. We used Glove 300 dimensional word embeddings trained on 6 Billion tokens with a 400,000 words vocabulary. For the Node2Vec training, we used balanced settings in the biased walks. Specifically, we used equal parameters for the breadth-first and the depth-first sampling. For the model training optimization, we employed Adam algorithm. In the following sections we explain the evaluation metrics, the comparison models, the main results analysis, and the model components ablation study.

\begin{table}[t]
\centering
\caption{The training settings.}
\label{tab:training}
\begin{tabular}{ll}
\hline
Settings&Value\\
\hline
Training Epochs&50\\
Early stopping epochs&5\\
Batch size&32\\
LSTM features in the hidden state&400\\
LSTM recurrent layers&2\\
Dropout probability&15\%\\
Learning rate&0.0008\\
Optimizer decay&35\%\\
Node representation padding&40\\
Node2Vec walk length&50\\
Node2Vec node walks&50\\
\hline
\end{tabular}
\end{table}

\subsection{Evaluation Metrics}

We evaluate our model on two commonly used metrics. First, the mean rank of the ground truth relation in the predicted relation probabilities set. Items in the set follow a descending order. Accordingly, the lower the mean rank the better the result. Since any pair of nodes may incur multiple correct relations, the rank of a correct relation should be decremented by the number of the valid relations that appeared on top of the correct one. This evaluation behaviour is called the filtered settings \cite{bordes2013translating} and is reported in our experiments along with the raw mean rank. The second evaluation metric is called Hits@1, that is the ratio of ground truth relations appeared as the first item in the ordered relations probability values. We also report the filtered settings for this metric.

\subsection{Comparison Models}

We perform a comparison with models that used the structural information and the textual information in the RP task. Although some models showed significant results in the same task, we excluded these results from our comparison because of the utilization of node information other than the node name, such as the description or the node entity type. For instance TaRP \cite{cui2021type} employed the node type information to achieve the reported performance. Nevertheless, most datasets do not have the node type information provided by default and that requires external knowledge bases with added load to retrieve and encode the type information. Additionally, TaRP did not generate embeddings for nodes but instead depended on existing models to do so. Similarly, the work in \cite{tagawa2019relation} used the entity description of long text to achieve similar results. In the following we show the models we consider in our comparison.

TransE \cite{bordes2013translating} was originally evaluated on the link prediction task. However, several works used the method for the RP task, and we use the reported rank and hits results in our comparison. TransE utilized the structural information to represent nodes. TransR \cite{lin2015learning} was evaluated on the link prediction and triple classification tasks. Similarly, other works reported TransR's performance in the RP task. TransR also utilized the structural information to represent nodes. KG-BERT \cite{yao2019kg} reported their relation prediction performance in the FB15K dataset. As described in Section \ref{sec:related_work}, the model's main design targeted the link prediction task. KG-BERT was also evaluated on the triple classification task. Shallom \cite{demir2021shallow} tackled the RP task and utilized neural networks embedding layers to represent nodes.

\begin{table}[t]
\centering
\caption{The results of our model and the comparison models on the FB15K dataset.}
\label{tab:results}
\begin{tabular*}{\linewidth}{@{\extracolsep{\fill}}lcccc}
\hline
Model&Mean Rank&Hits@1&Filtered Mean Rank&Filtered Hits@1\\
\hline
TransE \cite{bordes2013translating}  &-    &65   &2.5&84\\
TransR \cite{lin2015learning}  &-    &42   &2.1&91\\
KG-BERT \cite{yao2019kg} &1.69    &69   &1.25&\textbf{96}\\
Shallom \cite{demir2021shallow} &1.59 &73   &1.26&95\\
Ours    &\textbf{1.53}   &\textbf{74}   &\textbf{1.18}& 94\\
\hline
\end{tabular*}
\end{table}

\begin{table}[t]
\centering
\caption{The results of RPEST variants on the FB15K dataset.}
\label{tab:abalation}
\begin{tabular*}{\linewidth}{@{\extracolsep{\fill}}lccccc}
\hline
Model&Mean Rank&Hits@1&Filtered Mean Rank&Filtered Hits@1&Epoch Time\textsubscript{min}\\
\hline
RPEST\textsubscript{Glove} &1.63   &68   &1.26& \textbf{94}&\textbf{5}\\
RPEST\textsubscript{BERT} &1.70   &67   &1.31& 92&13\\
RPEST    &\textbf{1.53}   &\textbf{74}   &\textbf{1.18}& \textbf{94}&6\\
\hline
\end{tabular*}
\end{table}

\subsection{Main Results}

We show our main results in Table \ref{tab:results}. The results show superiority of our relation prediction model compared to the best results in other RP models. However, KG-BERT has equivalent filtered Hits@1 score. Nevertheless, our model beats the scores of the same model on the remaining metrics with good margin. The mean rank scores for TransE and TransR were not reported or found in any other study. We reason our model's superiority by the combination of the structural and textual details for every node. More precisely, we find the efficient utilization of the textual content leading the performance to this level as we show in the ablation study.

\subsection{Ablation Study}

To evaluate the effectiveness of our model's units, we created two variants to observe the model results after excluding some of its modules. Particularly, we created a variant that does not employ the nodes' structural information in the input. Therefore, we excluded the usage of Node2Vec in our model and we only kept Glove text encoder by creating a variant named RPEST\textsubscript{Glove}. Furthermore, we evaluated our choice of language models by replacing Glove with BERT language model in a variant named RPEST\textsubscript{BERT}. We show the evaluation metric scores of the different variants in Table \ref{tab:abalation}. The table also shows the epoch time in minutes for each variant, and we rely on the time values to emphasize the performance difference.

The lower scores of the RPEST\textsubscript{Glove} variant proofed the importance of employing the structural information in representing nodes. However, the variant had the fastest performance due to the reduced embeddings size after excluding Node2Vec vectors. On the other hand, the Glove variant scored better results compared to the RPEST\textsubscript{BERT} variant. We reason that by the efficient exploitation of the latent features in Glove's embeddings using the bi-directional LSTM layer along with the attention layer and the employment of the additional node type vector. The node type vector enhanced the relation direction detection in our model. Regarding the slowest performing variant, the large embeddings size led to the reported speed in BERT's variant.




\section{Conclusion} \label{sec:conclusion}

In this paper we explain that the relation prediction task has equal importance to the link prediction task. We also show that the usage of nodes' content information along with the structural information is important to have efficient graph representation. Accordingly, we propose a relation prediction model that utilizes a walks-based algorithm to retrieve the node structural information along with the utilization of pre-trained language models to represent the textual content of the nodes. We argue that masked language models are costly in terms of computational resources and pre-trained language models can achieve equivalent results with effective exploitation. Our model does not require the nodes' description and shows competitive results when compared to the state of the art models in the same task.






\bibliographystyle{splncs04}
\bibliography{springerkg}

\end{document}